# Automatic deductive coding in discourse analysis: an application of large language models in learning analytics

Lishan Zhang[a], Han Wu[b], Xiaoshan Huang[c], Tengfei Duan[b], Hanxiang Du[d]

[a]Department of Education, Beijing Institute of Technology, Beijing, China;
[b]Faculty of Artificial Intelligence in Education, Central China Normal University, Wuhan, China;
[c]Department of Educational and Counselling Psychology, McGill University, Montréal, Canada;
[d]Department of Computer Science, Western Washington University, Washington, United States

*Emails:lishan.zhang3@gmail.com, han.wu@mails.ccnu.edu.cn, xiaoshan.huang@mail.mcgill.ca, duan2237977549@mails.ccnu.edu.cn, duh2@wwu.edu*
*Lishan Zhang and Han Wu contribute equally to the study*

**ABSTRACT:** Deductive coding is a common discourse analysis method widely used by learning science and learning analytics researchers for understanding teaching and learning interactions. It often requires researchers to manually label all discourses to be analyzed according to a theoretically guided coding scheme, which is time-consuming and labor-intensive. The emergence of large language models such as GPT has opened a new avenue for automatic deductive coding to overcome the limitations of traditional deductive coding. To evaluate the usefulness of large language models in automatic deductive coding, we employed three different classification methods driven by different artificial intelligence technologies, including the traditional text classification method with text feature engineering, BERT-like pretrained language model and GPT-like pretrained large language model (LLM). We applied these methods to two different datasets and explored the potential of GPT and prompt engineering in automatic deductive coding. By analyzing and comparing the accuracy and Kappa values of these three classification methods, we found that GPT with prompt engineering outperformed the other two methods on both datasets with limited number of training samples. By providing detailed prompt structures, the reported work demonstrated how large language models can be used in the implementation of automatic deductive coding.

## 1. Introduction

Discourse analysis investigates the functional use of language to perform actions and construct identities, focusing on the meaning conveyed rather than the structural aspects or surface features of the language (Hjelm, 2021). Discourse analysis can be accomplished based on text, spoken language and recordings. In learning sciences, it is used to understand teaching and learning interactions in class or after-class tutoring (Rosé, 2017), which provides valuable insights for scaffolding and facilitating learning process(Wang, Yang, Wen, Koedinger, & Rosé, 2015; Meyer, 2023).

One common method for discourse analysis is deductive coding, which can transform qualitative analysis into quantitative analysis. In deductive coding, researchers first code each segment of discourses according to a predefined coding schema and then apply any statistical methods to analyze the coded results. This made discourse analysis easy to follow and more reproducible. However, the conduction of deductive coding is quite time-consuming because researchers need to human label all discourses to be analyzed according to theory-informed coding schema.

To overcome the disadvantage of deductive coding, researchers started using natural language processing technologies to automate the coding process, thereby saving precious research time (Rosé et al., 2008). This shift towards automated deductive coding not only addresses the time constraints but also enhances efficiency in data analysis. We consider such methods of using natural language processing to code discourse as automatic deductive coding.

This analysis technique facilitated mining the sentiments embedded in the discussion discourses. Researchers are able track the learning sentiments embedded in huge discussion data with the help of natural language processing technology. Additionally, beyond the immediate benefit of saving research time, developing and refining such technology serves as a foundational step towards the creation of dialog-based intelligent tutoring agents (Rosé & Ferschke, 2016).

From a technical perspective, automatic deductive coding shares similarities with automatic grading student's answers to open-response questions, which have been studied extensively (Zhang, Huang, Yang, Yu, & Zhuang, 2022). They both aim to classify students' plain text into several predefined classes. This is a classical text classification task in natural language processing. The classes can be either coding labels for deductive coding tasks or grades for auto-grading tasks. Automatic grading has been studied for decades. With the advent of advanced natural language processing methods such as LSTM and BERT, auto-grading accuracy has significantly improved. In contrast, while learning analytics researchers use text classification techniques in emotion and sentiment analysis, these techniques do not seem to have a widespread impact on traditional discourse analysis. Researchers still often need to manually code all text instead of referring to automatic coding or text classification techniques. This is probably because such text coding tasks often cannot provide enough training data.

The recent achievement in large language models, particularly GPT made by OpenAI, has opened up a new way for automatic deductive coding. GPT demonstrates remarkable performance on general classification tasks with few examples, known as few-shot learning, or even without any samples, which are called as zero-shot learning. Some recent studies have begun to explore the use of LLM like GPT for tasks related to qualitative discourse analysis, specifically focusing on deductive coding. However, while these studies have illustrated LLM's potential, limitations remain, especially concerning the LLM's advantages comparing to other machine learning methods and the different design strategies of prompts.

Unlike previous research that relied primarily on expert-developed codebooks, we explore the integration of fine-tuning techniques and retrieval-augmented generation to enhance the model's performance in deductive coding. By introducing these advanced methods, we aim to improve the adaptability of LLM in qualitative research, reducing reliance on rigid codebooks and providing more flexibility in coding diverse data sets. Moreover, our comparison across traditional text classification, BERT-like models, and GPT-based LLM offers new insights into the relative performance of these models in different qualitative data environments, highlighting areas where LLM can outperform of complement existing methods.

By conducting the exploration and classification methods comparison, the study aims for answering the following two research questions:
(1) How do the three classification methods perform with the given two data sets?
(2) How much improvements can we make by integrating the LLMs related techniques such as fine tuning and RAG?

The rest of the paper first briefly reviewed the existing works regarding sentiment analysis and auto-grading. These are the two fields where text classification algorithms have well proved their successes. Then we described two different data sets of our experiments. In the third, we introduced the three different text classification approaches with the emphasis in GPT approach. In the fourth, we reported the results with all the different settings. In the last, we concluded with remarks.

## 2. Related work

### 2.1. Emotion and sentiment analysis

In the field of learning analytics, text classification technology is often employed for the analysis of students' emotions in participatory learning processes. Participatory learning, different from rigid teacher-controlled instruction, underscores the active involvement of students in collaborative learning processes. The key to successful participatory learning is to create positive experiences that enhance children's cognitive engagement and awareness (Vartiainen, Tedre, & Valtonen, 2020). Emotions refers to a multi-componential construct of psychological subsystems, including affective, cognitive, motivational, expressive, and peripheral physiological processes (Pekrun, 2006). With the support of advanced technology, research on emotions and their role in cognitive processes within technology-rich learning environments has been gaining more attention (e.g., Huang, Huang, & Lajoie, 2022). In technology-supported learning contexts, emotions can be expressed through emotional tones in the form of verbal or textual output, which refers to the vocal expression or dialogue of emotion that conveys a student's affective states (Chang et al., 2023; Ishii, Reyes, & Kitayama, 2003). Emotions and sentiments interpreted from expressed emotional tones are associated with students' engagement, social interactions, and knowledge-sharing behaviors (Dang-Xuan et al., 2017; Näykki et al., 2014; Rapisarda, 2002), thereby shaping learners' overall learning experience. Automatically detecting learners' emotions form the foundation for understanding what keeps their learning experience positive and how to maintain such experiences. Therefore, it is important to automatically assess students' emotions and sentiments in real-time learning procedure with the assistance of machine learning and large language models.

Kastrati et al. (2020) devised a weakly supervised aspect-based sentiment analysis framework. Given student comments, they employed Convolutional Neural Network (CNN) for sentiment classification and Long Short Term Memory (LSTM) for aspect category recognition. Experiments were conducted using a substantial real-world education dataset, encompassing approximately 105K students' reviews gathered from Coursera, along with a dataset comprising 5,989 students' feedback in traditional classroom settings. CNN sentiment classification was applied for the binary classification of aspect sentiment, achieving 82.1% F1 score, while LSTM aspect category recognition was employed for identifying four aspect categories, achieving 86.3% F1 score. Dahiya, Mohta and Jain (2020) performed sentiment and emotion analysis on textual messages with emoticons. Employing a CNN model, they trained a classifier to categorize 29,939 unique statements which are acquired from Kaggle into six distinct emotions and gave an average accuracy of 72.9%. Klünder, Horstmann and Karras (2020) integrate sentiment analysis with tradition natural language processing techniques to automate sentiment classification in text-based communication. They utilized three machine learning methods—random forest, Support Vector Machine (SVM), and Naive Bayes—to categorize each text segment as positive, neutral, or negative. The efficacy of this approach is substantiated through an industrial case study in software development. The case study comprises 1,947 messages extracted from a group chat within the Zulip communication tool, encompassing a total of 7,070 sentences. The ultimate classification results, reaching an accuracy of 62.97%, exhibit a level of effectiveness comparable to human ratings (Klünder et al., 2020).

Generally, datasets used for sentiment analysis are large and need a lot of human labeling for training. Even traditional methods such as random forest and SVM can sometimes satisfy the requirements. However, certain limitations may exist, such as the inability to integrate contextual information from the dialogue for assessment.

## 2.2. Automatic grading for open-response questions

The grading of open-response questions is a critical aspect of educational assessments, requiring a balance between subjective evaluation and the need for efficient, scalable, and consistent grading methods. Grading for open-response questions involves the assessment of students' answers to questions that require free-form responses, as opposed to multiple-choice or other closed-ended formats. This context often demands more flexible and subjective grading methods in educational and examination settings. So traditional approach involves manual grading by educators or domain experts. To save repetitive human works and support personalized learning (Erickson & Botelho, 2021), recent researches have focused on the development of automatic grading methods with machine learning and natural language processing technologies.

Erickson et al. (2020) employed tree-based machine learning approaches, including random forest and XGBoost, as well as deep learning methodologies such as LSTM and the Rasch Model, to assess and analyze open-response questions in mathematics. The dataset utilized in their study comprised a total of 141,612 student responses to 2,042 unique problems from 25,069 students. Wilson et al (2022) conducted a comparative analysis employing three machine learning models including logistic regression, random forest, and K nearest neighbors. These algorithms are used to classify 2,450 student responses to open-ended questions in the Physical Measurement Questionnaire into four classes. Zhang et al (2022) utilized the continuous bag-of-words model (CBOW) to integrate the domain-general and domain-specific information in the process of feature engineering. Then they built the classifier model using LSTM and evaluated it with 7 reading comprehension questions with over 16,000 labeled student answers.

Compared with other traditional automatic grading models, their proposed model significantly improved the automatic grading performance on semi-open-ended questions.

**2.3. LLM-based models for deductive coding**

Building on the advancements in emotion and sentiment analysis, as well as the development of automatic grading systems for open-response questions, there has been growing interest in applying LLM to other complex tasks such as deductive coding in qualitative research. To offer new possibilities for handling large-scale qualitative data sets more efficiently, researchers have begun exploring how these models can assist or automate parts of this process. Xiao et al (2023) find that combing LLM with expert-drafted codebooks achieves fair to substantial agreements with expert-coded results in deductive coding. Their study utilized an expert codebook to construct prompts, and although it provided transparency and explicit control, it limited the performance of the model. Tai et al (2023) proposed a methodology using LLM to support traditional deductive coding in qualitative research. They compared the performance of a large language model with traditional human coding in identifying five conceptual codes in three narrative texts, providing a systematic and reliable platform for code identification and offering a means of avoiding analysis misalignment. They also pointed that there is a lack of validated research examining the use of LLM in qualitative analysis. Chew et al (2023) provide a holistic approach for performing deductive coding with LLM, and aims to assess the effectiveness of GPT-3.5 across a range of deductive coding tasks through an in-depth case study and empirical evaluation on four publicly available datasets. The study noted several limitations, including the need for extensive prompt engineering and the assessment of coding performance across a wide variety of LLM. Hou et al (2024) explored the use of LLM to assist in deductive coding of social annotation data, achieving fair to substantial agreement with human raters in context-independent dimensions and moderate agreement in context-dependent dimensions. They claimed that there were still some challenges of including original text in the prompt engineering or fine-tuning models for context-dependent dimensions.

In summary, existing works regarding text classification adoption in education mainly focus on sentiment analysis in participatory learning and automatic grading on student answers. Both of these works required considerable large data set for training the supervised machine learning models, so that satisfactory results can be achieved. Only a few studies have attempted to use LLM for deductive coding in content analysis. Our study aims to further verify that generalist foundation models such as GPT and their combination with other technologies can be helpful in deductive coding tasks with few labeled items. In particular, we took two different datasets to conduct our experiments.

# 3. Datasets

To conduct our study and evaluate the three automatic coding techniques, we used two datasets that were obtained from a Chinese poetry appreciation literary course at a university in central China. The course was open for science and engineering students, aiming to improve their literacy ability and cultivate their aesthetic ability. The course adopted a face-to-face collaborative teaching model and the classroom students were divided into groups. Some pre-class or in-class tasks require students to complete through a collaborative annotation platform developed by our research team. Before class, the teacher assigned reading tasks, requiring students to read learning materials on the platform and highlight and annotate the materials. This formed the annotation dataset. During class, students used the chat section of the platform to interact within groups based on in-class tasks and collaborate to solve problems. This formed the discussion dataset. We described the two datasets and the coding schema in detail below.

Each dataset was split into training and testing sets using an 8:2 ratio. The training set was to train the machine learning models or tune the prompts. The testing set was to evaluate the performance of different automatic coding techniques. In this way, all the automatic coding models used the exact same dataset for model calibration and evaluation. Note that the GPT approach only used several items in the training set and the other machine learning approaches used the entire training set. However, such settings formed fair comparisons for all the models.

**3.1. The Annotation dataset**

In the fall semester of 2021, we collected a dataset comprising 607 student annotations. This dataset was divided into a training set containing 484 pieces of data and a testing set containing 123 pieces. Seventy-three students participated in the course, representing diverse majors such as artificial intelligence, history, psychology, physics, etc.

Students were asked to read and annotate a Chinese article titled "The Image of Plum in Ancient Chinese Literature" on a reading annotation system developed by our research team. The annotation reading interface of the system is shown in Figure 1.

**Figure 1**

*The annotation reading interface of the platform*

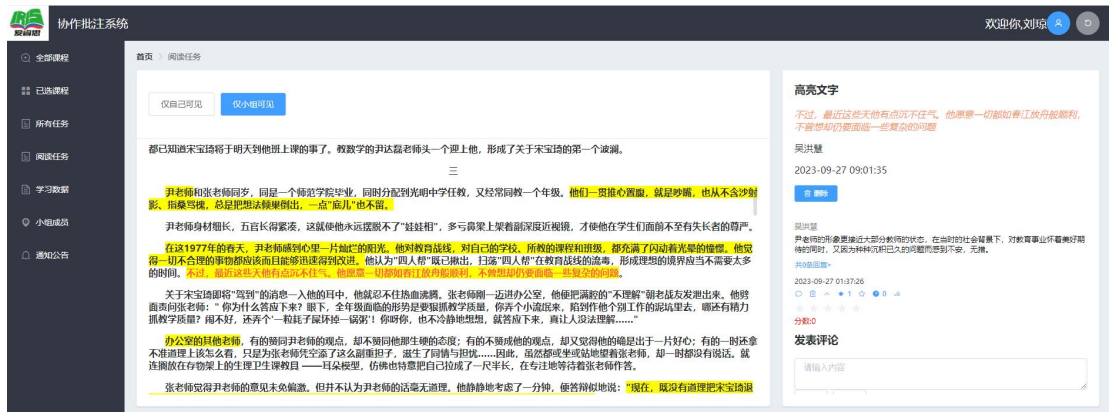

After collecting the data, two researchers manually analyzed and coded the annotation data according to the cognitive engagement classification scheme adapted from Chi and Wylie (2014). After randomly selecting 200 items from the annotation data, the level of agreement between the two researchers was measured at 0.752 Kappa, indicating good reliability. Cognitive engagement of the annotation data was analyzed manually based on the coding scheme of Table 1.

**Table 1**

*The coding scheme of annotation data*

| Code | Behaviors | Descriptions | Exemplar |
| --- | --- | --- | --- |
| A | Copy | Highlight and directly/selectively copy ideas from material. | "During the Six Dynasties period, plum blossoms served as symbols of friendship, love, and hometown sentiments, yet they had not yet freed themselves from the metaphorical expressions of comparison." |
| C1 | Construction | Make inferences, generalizations or summaries based on the highlighted content | "I believe this is inseparable from the characteristics of plum blossoms. Blooming in the harsh winter, preceding the myriad flowers, they stand alone in ushering in spring. This resilience in blooming independently in the cold resonates with many poets who have faced adversity in their careers and lives, yet refuse to conform to worldly impurities." |
| C2 | Integration | Highlight, and integrate other information in the material or other materials for comparison and connection, etc. | "In the ancient poems I've studied before, there were "red beans" symbolizing love and "broken willows" conveying homesickness. Now, I've come to realize that "plum blossoms" can also represent friendship, love, and hometown sentiments." |

## 3.2. The Discussion dataset

The discussion dataset, comprising a total of 404 pieces of discussion data, was gathered in the fall semester of 2022. This dataset was subsequently divided into a training set containing 320 pieces of data and a testing set containing 84 pieces. Seventy-two students, voluntarily enrolled in the course, participated in this dataset. The students had an average age of 20 and came from various non-literary majors. The students were grouped based on the Kolb Learning Style Questionnaire(Manolis, Burns, Assudani, & Chinta, 2013), resulting in a total of 10 groups with 6-7 students in each.

Students worked in groups to select one of the five poets and analyzed their representative works of chrysanthemum poems through text communication using the same system. The discussion interface of the system is shown in Figure 2.

**Figure 2**

*The discussion interface of the platform*

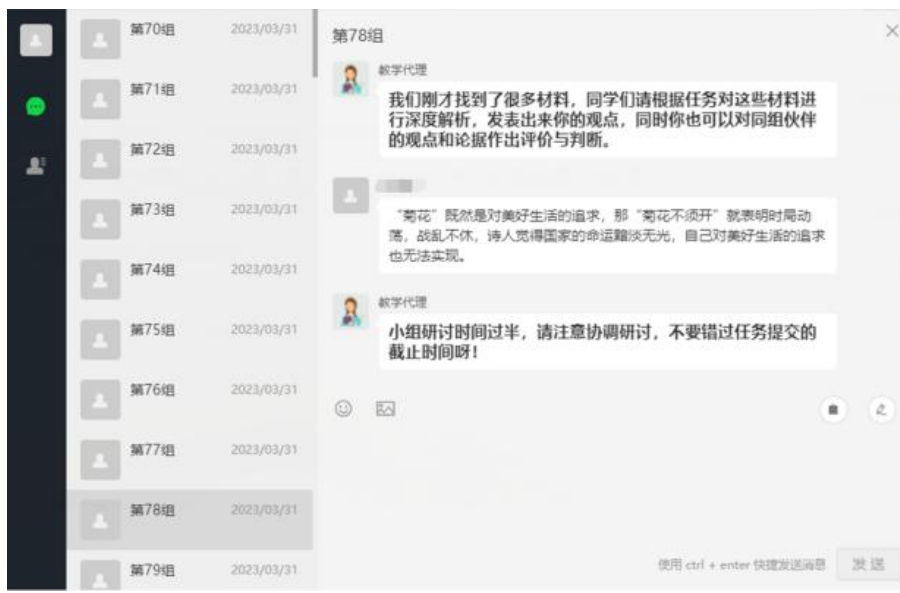

Following data collection, two researchers conducted manual coding and analysis of the discussion data. The coding scheme, a modified iteration of Chi's cognitive engagement framework (Chi & Wylie, 2014), was tailored to students' discussion tasks and learning characteristics. In the initial phase, two researchers independently selected data from two groups for coding, amounting to a total of 84 instances, achieving an inter-rater reliability of 0.69. Following this, the researchers refined the coding framework and deliberated on the initially coded data. Through negotiation and consensus-building, they proceeded with a second round of pre-coding. In this subsequent round, two groups, comprising a total of 69 instances, were once again randomly chosen, resulting in an enhanced inter-rater reliability of 0.80. The cognitive engagement coding framework employed by researchers during the manual analysis of the discussion data is detailed in Table 2.

**Table 2**

*The coding scheme of discussion data*

| Code | Descriptions | Exemplar |
|---|---|---|
| M | Assigning, coordinating, and supervising tasks. | "Sure, let's start with the translation, shall we?" |
| P | Simply copying the original text or online information, or summarizing others' opinions. | "The term "wild jackals offering to the moon" refers to the first solar term during the Frost's Descent period, where jackals ceremoniously present their prey before consuming it. Ancient people believed this behavior to be a form of wild jackals performing a ritual to the moon." |
| A | Expressing agreement. | "Yeah, you're right." |
|   | Expressing opposition. | "Hahahahahahaha no." |
| C | Raising cognitive doubts or questions. | "The youth has already departed - Does this imply that the poet no longer possesses the youthful vigor and no longer aspires to the court?" |
|   | Independently proposing a new and original point of view. | "So, the first sentence roughly means that the time of youth has already passed and won't come back." |
| I | Agreeing with a partner's perspective and providing additional explanations. | "Sure, actually, the chrysanthemums at this time gave him a lot of inspiration and temporarily lifted him from the feelings of loneliness." |
|   | Opposing a partner's viewpoint and providing additional explanations. | "But emotions shouldn't be, right? He's feeling down and lonely right now." |
|   | Responding to and supplementing questions and doubts from peers. | "No way, didn't he later still have confidence in joining the court?" |

The three automatic coding techniques were evaluated by following the procedure shown in Figure 3. The manually coded results were divided into training and testing datasets. The three automatic classification methods, including the traditional text classification method, BERT-like pretrained language model and GPT-like pretrained language, were applied to classify the annotation and discussion data. The classification methods are detail described in the next section.

**Figure 3**

*The experiment procedures*

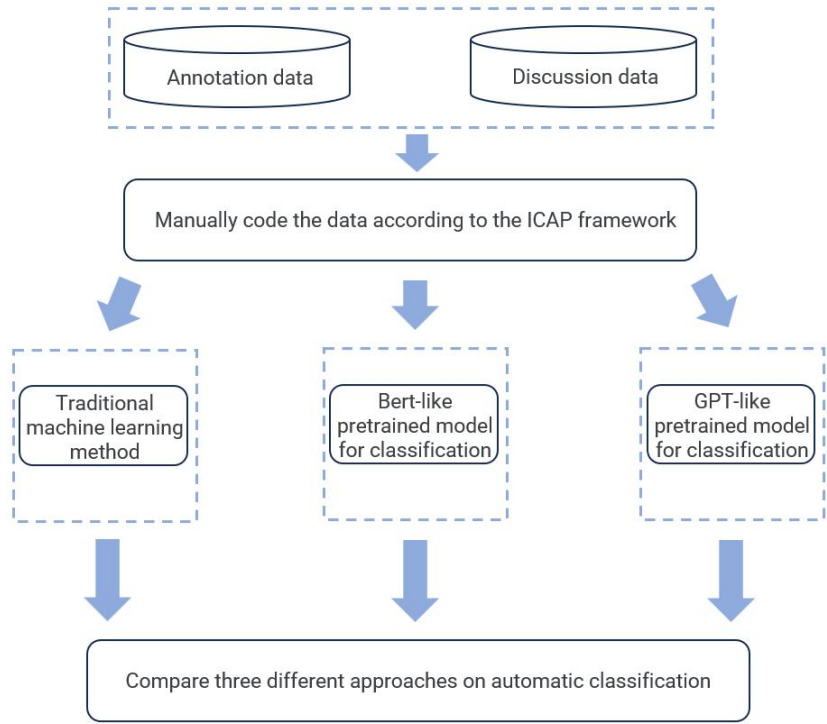

## 4. Methods

This section briefly introduced the three different approaches of text classification for automatic deductive coding. In this study, we considered two different types of discourses. One is students reading annotation, the other is student discussion dialog. Since that the purpose of the comparison is to explore the power of large language models, we respectively selected one representative model for traditional machine learning and BERT-like methods. The rest of the section described these two models and how we used GPT on deductive discourse coding.

### 4.1. Traditional machine learning method

We selected Random Forest (RF) as the representative model for traditional machine learning method, because RF has proved to be well performed in many related tasks (Schonlau & Zou, 2020). Figure 4 illustrates the process of using traditional machine learning method for classification. The very first step is to conduct word segmentation to transform each student sentence into a set of words. We used jieba library to perform this step. Note that such a transformation would lose the sequential information of the original sentence as well as the grammatical structures (Qader, Ameen, & Ahmed, 2019). After the sentences were split into sets of words, the frequency of each word can be calculated and the corresponding frequencies of the words used as the inputs of Random Forest. This kind of feature engineering approach is called as Bag-of-Words (BoW). Random Forest is essentially a group of decision trees. Each decision node of the decision trees was comprised of word frequency and a threshold. We used RandomForestClassifier in scikit-learn to perform the implementation. There were two hyper-parameters of this algorithm, which were estimator and random feed. Estimator defines the upper bound of the number of decision trees in the forest and the random feed decides initial values of the parameters. The estimators was set to 100 and the random feed was set to 42.

**Figure 4**

*Process of Random Forest classification method*

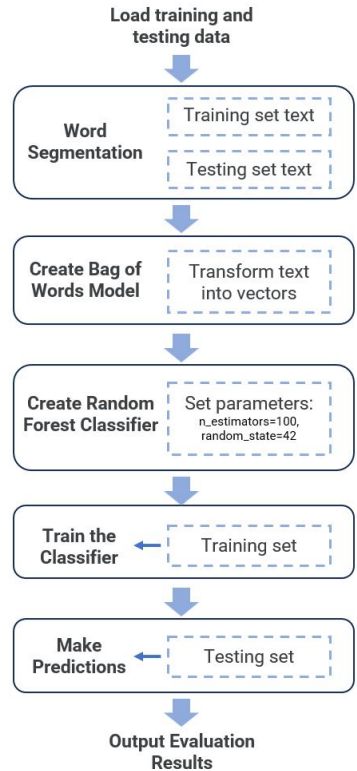

### 4.2. BERT-like pretrained model for classification

For BERT-like pretrained models, we used RoBERTa, an advanced version of BERT, enhanced through training on a larger corpus (Shaheen, Wohlgenannt, & Filtz, 2020). Notably, Qasim et al (2022) demonstrated the superior performance of the RoBERTa model in various classification tasks compared to other pre-trained language models. However, they also highlighted that RoBERTa exhibits stronger linguistic bias.

Figure 5 shows the process of using RoBERTa for classification. The initial step involves loading the dataset and tokenizing it using RoBERTa's tokenizer. This process converts the text into numerical representations suitable for machine learning. In specific, we utilized the chinese_roberta_wwm_ext_pytorch pre-trained model. Different pooling strategies, including CLS pooling, mean pooling, and max pooling, are implemented to extract features from the hidden states of the RoBERTa model. These strategies contribute to capturing essential information from the input sequences. A linear classifier head is incorporated into the model to translate the extracted features into target categories. The model is trained using the Adam optimizer with a learning rate of 2e-5 and a weight decay of 1e-4. The training is conducted over 5 epochs. A batch size of 16 is employed during the training process. The maximum sequence length for tokenization is set to 200. Sentences exceeding this length are truncated, while shorter sentences are padded with spaces to ensure uniform input dimensions.

**Figure 5**

*Process of RoBERTa classification method*

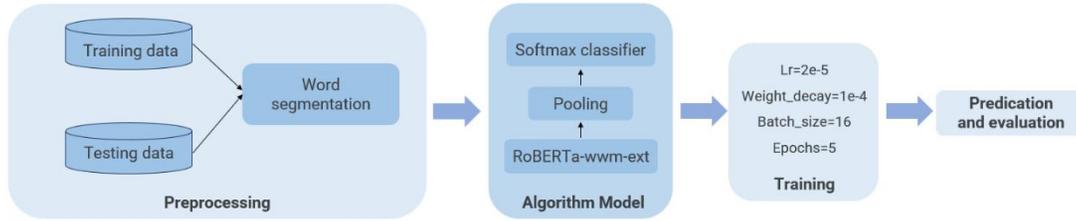

### 4.3. GPT-like pretrained model for classification

Despite the proliferation of various generative artificial intelligence models, GPT developed by OpenAI stands out as the most representative and proficient in multiple aspects. In many cases, GPT can perform well with only the instructions given in the prompts. To better instruct GPT, researchers often need to go through a process called as prompt engineering. Prompt engineering is the means by which GPT are programmed using prompts (White et al., 2023). Few-shot learning and chain of thought (CoT) are two well known and useful techniques in prompt engineering. Few-shot learning aims to enable models to learn and adapt to new tasks from a limited number of samples (Brown et al., 2020), while CoT achieves this by breaking texts into continuous segments and utilizing them as model inputs (Wei et al., 2022). Recently, another technique is called as retrieval-augmented generation (RAG) is has emerged that can integrate domain knowledge into general large language models, thereby improving the quality and effectiveness of the generated results(Gao et al., 2023). Besides prompting engineering, fine-tuning is another method of calibrating the large language models to fit the downstream tasks (Touvron et al., 2023). But this method required much more labeled samples for training the models.

In this study, we considered all the techniques mentioned above, including few-shot learning, CoT, RAG and fine-tuned. In addition, we combined traditional natural language processing techniques with GPT for further improvement.

In terms of the foundation model, this study mainly employed GPT-4, recognized as one of the most powerful generative AI models. The fine-tuned foundation model was GPT 3.5 because OpenAI did not support fine-tuning GPT-4. To optimize the prompts, we used GPT4's playground for prompt formulation and refinement. When we were satisfied in the playground, we used the final prompt we got to fabricate the backend code, which went through all the discourse data and called OpenAI's API to process it. We used four different settings in the GPT approach to explore the usefulness of the techniques. Because the dataset we used was in Chinese, all prompts were also written in Chinese in our actual program no matter the setting. We translated them into English for ease of reading.

#### 4.3.1. Prompt only

In this method, we exclusively employed prompts and GPT-4 for deductive coding. During the prompt tuning process in the playground, we drafted a prompt based on the dataset characteristics. Through iterative refinement, our final prompt consisted of the five parts below (a full version sample prompt is attached in the appendix):
- 【Introduction to the Course Background】
- 【Issuance of Instructions】
- 【Detailed Introduction to Encoding Rules】
- 【Output Structure and Examples】
- 【Input data】

At first, the prompt introduced the background of the course where the student discourse was generated. Then the prompt described what GPT needed to do (i.e. conduction of deductive coding). In the third, the prompt provided the detailed codebook. In the following, the prompt described the desired format of the output. Several coding examples were also provided here as the implementation of few-shot learning techniques. In specific, we manually selected three representative student discourse examples and told GPT the corresponding human labels. In the last part of the prompt, we gave the input data that needs to be coded. This method was only used for the annotation dataset. So, the input data comprised the annotation along with its highlighting text.

Regarding GPT settings, we used the gpt-4-1106-preview version with a temperature value of 0, a maximum text length of 4096, frequency_penalty set to 0, and presence_penalty set to 0.

*4.3.2. Finetuning*

We performed fine-tuning based on the GPT-3.5-Turbo model. The basic idea of fine-tuning is to use many labeled examples as the training data to calibrate the foundation model, so that the fine-tuned model can adapt to the specific downstream task, which is deductive coding in our case. As the documents of GPT suggested, around 100 trainning samples would be sufficient for the fine-tuning task of GPT. This method was applied to both annotation and discussion datasets.

For the annotation dataset, we used the same prompts as those in the previous setting (i.e. prompt only). We used 90 entries to build fine-tuned model and evaluated its performance. The training epoch was set to 3.

For the discussion dataset, because we had 10 sets of dialogs generated by 10 groups of students with 404 dialog turns in total, we asked GPT to consider the dialog turns independently without the context and did fine-tuning. We used 100 entries to build fine-tuned model. The training epoch was set to 3 as well.

*4.3.3. Prompt + traditional NLP*

As Do et al (2024) suggested, the overall performance can be improved by integrating prompt based LLMs and traditional NLP technologies. So, we did such combinations in our automatic deductive coding as well. In specific, we built a reference database including the original reading material and the corresponding online reference information. During the process of automatic coding, instead of handling over the entire task to GPT, we did similar sentence checking at first. It means that for each input sentence of students' annotations, we wrote program to find out the most similar sentence in the reference database. The similarity between two sentences is defined as the follows:

$$similarity_{cos}(A, B) = \frac{\sum_{i=1}^{n} A_i \times B_i}{\sqrt{\sum_{i=1}^{n} (A_i)^2} \times \sqrt{\sum_{i=1}^{n} (B_i)^2}} \#(1)$$

If the similarity score between a student's annotation and the most similar sentence in the reference database exceeded a threshold, the program coded the annotation as "A", otherwise, the program asked GPT to make further judgement. The structure of the prompt for the further judgement remained the same, except for the absence of the introduction to the code "A".

*4.3.4. Prompt with context knowledge + traditional NLP*

At last, we further included context knowledge in the prompt. We used different strategies for the two different datasets. We used RAG with the reference database mentioned in the previous section for the annotation dataset. We injected the full dialog context for the discussion dataset. We respectively described the two strategies in the following.

For the annotation dataset, we retrieved two relevant sentences of the input annotation from the reference database based on their sentence similarities. We used the retrieved information to augment the generated results of GPT, in which retrieval augmented generation (RAG) was implemented. Note that the traditional NLP technique was used again in the calculation of sentence similarity. Different from the prompts introduced previously, prompts here were constructed dynamically based on the reference sentences. The contents of the prompts were also slightly different because we asked GPT to break the deductive coding process into two steps inspired by CoT. In specific, GPT needed to first consider the relation of the student's annotation to the two reference sentences, then make the coding decision. Figure 6 illustrated how traditional NLP and context knowledge were integrated in the workflow of the automatic deductive coding with GPT. Traditional NLP was integrated mainly through similarity calculation and context knowledge was integrated through RAG. The resulting framework of the classification prompt for this experiment is outlined below:

- 【Introduction to the Course Background】
- 【Issuance of Instructions】
- 【Detailed Introduction to Encoding Rules】
- 【Output Structure and Examples】
- Comment
- Highlight
- Reference1
- Reference2

**Figure 6**

*The workflow of automatic deductive coding for annotation data*

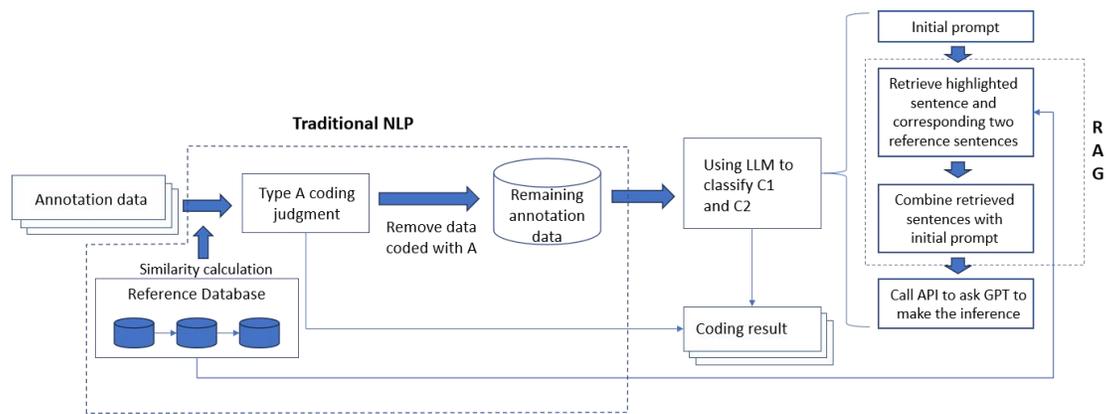

For the discussion dataset, we included the entire dialog of the discussion group as the context knowledge and asked GPT to code each dialog turn. GPT was instructed to consider the surrounding dialog turns while determining the encoding categories. The framework of the classification prompt for this dataset is as follows:

- 【Introduction to the Course Background】
- 【Issuance of Instructions】
- 【Detailed Introduction to Encoding Rules】
- 【Output Structure and Examples】
- Student dialogs

## 4.4. Evaluation metrics

In this section, we described the evaluation metrics employed to assess the performance of the automatic deductive coding approaches. We used accuracy, precision, recall, F1 score, and Cohen's kappa to evaluate and compare the approaches. The first four metrics are widely used in computational-related journals and the last one is used widely in psychology journals.

Accuracy represents the ratio of correctly predicted instances to the total instances in the dataset. It provides an overall measure of the model's correctness. Precision is the ratio of correctly predicted positive observations to the total predicted positives. It measures the accuracy of positive predictions. Recall, also known as sensitivity or true positive rate, is the ratio of correctly predicted positive observations to the actual positives in the dataset. It assesses the model's ability to capture all relevant instances. The F1 score is the harmonic mean of precision and recall. It provides a balanced measure that considers both false positives and false negatives. Cohen's Kappa is a statistic that

*12*

measures the agreement between the predicted and actual classifications, considering the possibility of the agreement occurring by chance. It corrects for the chance agreement inherent in accuracy. High accuracy, precision, recall, and F1 score values all indicate effective classification performance. Kappa values indicate the amount of agreement between the predicted results and the ground truth. Values close to 1 indicate substantial agreement, while 0 suggests no agreement. This comprehensive set of evaluation metrics enables a thorough analysis of the classification methods, shedding light on their efficiency in handling annotation and discussion datasets.

## 5. Results and discussion

In this section, we reported the results of the three different approaches for automatic deductive coding and make a discussion on the results. For each approach, we reported the coding results of annotation and discussion datasets respectively. We made a summary of the three approaches by the end of this section. Among the three approaches, we focused on the GPT-based one.

### 5.1. Traditional machine learning (random forest)

For the annotation dataset, the random forest classification method achieved an overall accuracy of 0.56. This accuracy underscores a moderate yet promising success in accurately assigning coded annotations to their respective categories. The Kappa value was found to be 0.28. We calculated precision, recall, and F1-score for each code. The results were reported in Table 3.

Table 3

*Performance Metrics for Annotation Data*

|  | precision | recall | f1-score | support |
|---|---|---|---|---|
| A | 0.56 | 0.33 | 0.42 | 30 |
| C1 | 0.55 | 0.89 | 0.68 | 53 |
| C2 | 0.63 | 0.30 | 0.41 | 40 |
| Macro avg | 0.58 | 0.51 | 0.50 | 123 |
| Weighted avg | 0.58 | 0.56 | 0.53 | 123 |

As for the discussion dataset, the random forest classification method exhibited a lower overall classification accuracy of 0.48. The Kappa value for the discussion data was 0.32, indicating slight agreement beyond chance. This discrepancy in accuracy between the two datasets emphasizes the method's varying performance when applied to different types of data. Similarly, precision, recall and F1-score were reported in Table 4.

Table 4

*Performance Metrics for Discussion Data*

|  | precision | recall | f1-score | support |
|---|---|---|---|---|
| M | 1.00 | 0.62 | 0.76 | 13 |
| P | 1.00 | 0.25 | 0.40 | 20 |
| A | 0.32 | 1.00 | 0.49 | 12 |
| C | 0.45 | 0.54 | 0.49 | 28 |
| I | 0.00 | 0.00 | 0.00 | 11 |
| Macro avg | 0.56 | 0.48 | 0.43 | 84 |
| Weighted avg | 0.59 | 0.48 | 0.45 | 84 |

The results clearly showed that the random forest classifier did not provide good performance in general and had quite different performance on different coding categories. This approach exhibited an especially bad performance in the Interaction (I) coding category in the discussion dataset. This was too surprising because this approach simply treated all discourses as bags of words and made deductive coding decisions solely based on the frequencies of occurrences of these words.

### 5.2. BERT-based (RoBERTa)

For the annotation dataset, the overall accuracy was determined to be 0.59, with a Kappa value of 0.36. Precision, recall, F1-score for each coding category was reported in Table 5.

**Table 5**

*Annotation classification metrics using RoBERTa*

|   | precision | recall | f1-score | support |
|---|---|---|---|---|
| A | 0.64 | 0.47 | 0.54 | 30 |
| C1 | 0.66 | 0.75 | 0.70 | 53 |
| C2 | 0.47 | 0.47 | 0.48 | 40 |
| Macro avg | 0.59 | 0.57 | 0.57 | 123 |
| Weighted avg | 0.59 | 0.59 | 0.59 | 123 |

For the discussion dataset, the automatic coding results indicate an overall accuracy of 0.67, accompanied by a Kappa value of 0.54, showcasing a robust performance. The class-specific coding results are presented in Table 6.

**Table 6**

*Discussion classification metrics using RoBERTa*

|   | precision | recall | f1-score | support |
|---|---|---|---|---|
| M | 0.00 | 0.00 | 0.00 | 13 |
| P | 0.90 | 0.95 | 0.93 | 20 |
| A | 0.92 | 0.92 | 0.92 | 12 |
| C | 0.51 | 0.93 | 0.66 | 28 |
| I | 0.00 | 0.00 | 0.00 | 11 |
| Macro avg | 0.47 | 0.56 | 0.50 | 84 |
| Weighted avg | 0.52 | 0.67 | 0.57 | 84 |

As a more advanced NLP technique, RoBERTa demonstrated improved performance in both annotation and discussion datasets. However, this approach seemed to produce biased results more easily. For example, the trained automatic coder did not identify any M or I code. Probably because its computational model was complex and easy to be overfitted by a small amount of data.

### 5.3. GPT-based

For the annotation dataset, we ran four experiments with the automatic coding methods including prompt-only, fine-tuning, prompt +NLP, and prompt with context + NLP. We reported the four results in the following.

**Results of prompt-only method.** The overall accuracy was quite low, at 0.37. Given that it was essentially a three-fold classification problem, the performance was just a little bit better than chance. The Kappa value was only 0.005. So, using prompt alone was not enough for accomplishing our deductive coding task. Table 7 provides more details of the results for each coding category.

**Table 7**

*Annotation classification metrics of prompt only experiment*

|   | precision | recall | f1-score | support |
|---|---|---|---|---|
| A | 0.29 | 0.17 | 0.21 | 30 |
| C1 | 0.42 | 0.45 | 0.44 | 53 |
| C2 | 0.33 | 0.40 | 0.36 | 40 |
| Macro avg | 0.35 | 0.34 | 0.34 | 123 |
| Weighted avg | 0.36 | 0.37 | 0.36 | 123 |

**Results of fine-tuning method.** The performance of automatic deductive coding is significantly improved when the model after finetuned, even with a weaker GPT-3.5-Turbo version. The accuracy was improved to 0.54 with a Kappa of 0.28. The precision, recall and F1-score for each coding category were detailed in Table 8.

Table 8

*Results of the fine-tuning model for annotation classification of prompt only*

|              | precision | recall | f1-score | support |
|--------------|-----------|--------|----------|---------|
| A            | 0.61      | 0.67   | 0.63     | 30      |
| C1           | 0.59      | 0.64   | 0.61     | 53      |
| C2           | 0.38      | 0.30   | 0.33     | 40      |
| Macro avg    | 0.52      | 0.54   | 0.53     | 123     |
| Weighted avg | 0.52      | 0.54   | 0.53     | 123     |

**Results of prompt+NLP method.** When the active coding category was identified by the calculation of sentence similarities, the overall performance of prompt + NLP was improved compared with the prompt-only method. The overall accuracy was 0.46 and the Kappa was 0.164. Table 9 provides the performance details for each coding category. We also combined NLP technique here with the fine-tuned model. In a result, the overall accuracy was 0.63 and the Kappa was 0.43.

Table 9

*Annotation classification metrics with NLP method*

|              | precision | recall | f1-score | support |
|--------------|-----------|--------|----------|---------|
| A            | 0.85      | 0.73   | 0.79     | 30      |
| C1           | 0.45      | 0.49   | 0.47     | 53      |
| C2           | 0.23      | 0.23   | 0.23     | 40      |
| Macro avg    | 0.51      | 0.48   | 0.49     | 123     |
| Weighted avg | 0.47      | 0.46   | 0.47     | 123     |

**Results of prompt with context + NLP method.** The performance of automatic deductive coding achieved the best result when the context and NLP were both involved. The overall accuracy was 0.71 and the Kappa was 0.54. Detailed performance for each coding category was presented in Table 10.

Table 10

*Annotation classification metrics using GPT-4*

|              | precision | recall | f1-score | support |
|--------------|-----------|--------|----------|---------|
| A            | 0.85      | 0.73   | 0.79     | 30      |
| C1           | 0.68      | 0.83   | 0.75     | 53      |
| C2           | 0.66      | 0.53   | 0.58     | 40      |
| Macro avg    | 0.73      | 0.70   | 0.70     | 123     |
| Weighted avg | 0.71      | 0.71   | 0.70     | 123     |

For the discussion dataset, we ran two experiments with the automatic coding methods including fine-tuning and prompt with context + NLP.

**Results of fine-tuning method.** Remind that the fine-tuned model considered each turn in the dialogs independently. Without such context information, the corresponding automatic coding performance can still achieve an accuracy at 0.73 and the Kappa was 0.64. The detailed results of precision, recall and F1 score for each coding category were described in Table 11.

Table 11

*Results of the fine-tuning model for five-class classification of discussion data*

|              | precision | recall | f1-score | support |
|--------------|-----------|--------|----------|---------|
| M            | 0.67      | 0.92   | 0.77     | 13      |
| P            | 0.90      | 0.90   | 0.90     | 20      |
| A            | 0.77      | 0.83   | 0.80     | 12      |
| C            | 0.64      | 0.75   | 0.69     | 28      |
| I            | 1.00      | 0.00   | 0.00     | 11      |
| Macro avg    | 0.79      | 0.68   | 0.63     | 84      |
| Weighted avg | 0.77      | 0.73   | 0.68     | 84      |

**Results of prompt with context + NLP.** When the entire dialog was included the prompt, the accuracy of automatic deductive coding reached 0.77, with a Kappa value of 0.72. This overall performance is acceptable for inter-agreement. The detailed results for each coding category is presented in Table 12.

**Table 12**

*Discussion classification metrics using GPT4*

|              | precision | recall | f1-score | support |
|--------------|-----------|--------|----------|---------|
| M            | 1.00      | 0.62   | 0.76     | 13      |
| P            | 0.83      | 0.95   | 0.88     | 20      |
| A            | 0.80      | 1.00   | 0.89     | 12      |
| C            | 0.67      | 0.86   | 0.75     | 28      |
| I            | 1.00      | 0.18   | 0.31     | 11      |
| Macro avg    | 0.86      | 0.72   | 0.72     | 84      |
| Weighted avg | 0.82      | 0.77   | 0.75     | 84      |

Based on the reported results above, the same method had quite different performances on the two different datasets. It is probably because that writing annotation itself is more complex than discussing to solve in-class tasks within groups. Students usually do not think too much before writing down their opinions and can even have many casual talks during discussions. In addition, contextual information in discussion is easy to obtain. It is the discourses of the sibling turns of the dialogs. In contrast, while students are making reading annotations, students need to first select and highlight one or two sentences that they think are interesting in the reading material. Then, the students may integrate the highlighted information with their own thoughts to make the annotations. Some students may even web search for outside knowledge to make annotations of better qualities. Indeed, some students tend to simply copy and paste the information they searched. We found out we could easily identify these cases via sentence similarity calculation, which is a widely used NLP technique.

Due to the nature of the complexity of annotation coding, we ran four experiments with different methods on the annotation dataset. We summarized and illustrated the performances of four different settings in Figure 7. We can notice that fine-tuning, the combination of NLP, and providing context information all benefit the coding performance. So, it seems necessary to integrate all suitable techniques to dynamic construct prompts for automatic deductive coding instead of only relying on LLMs with static prompts.

**Figure 7**

*Comparison of Classification Metrics for Annotation Data across All Methods*

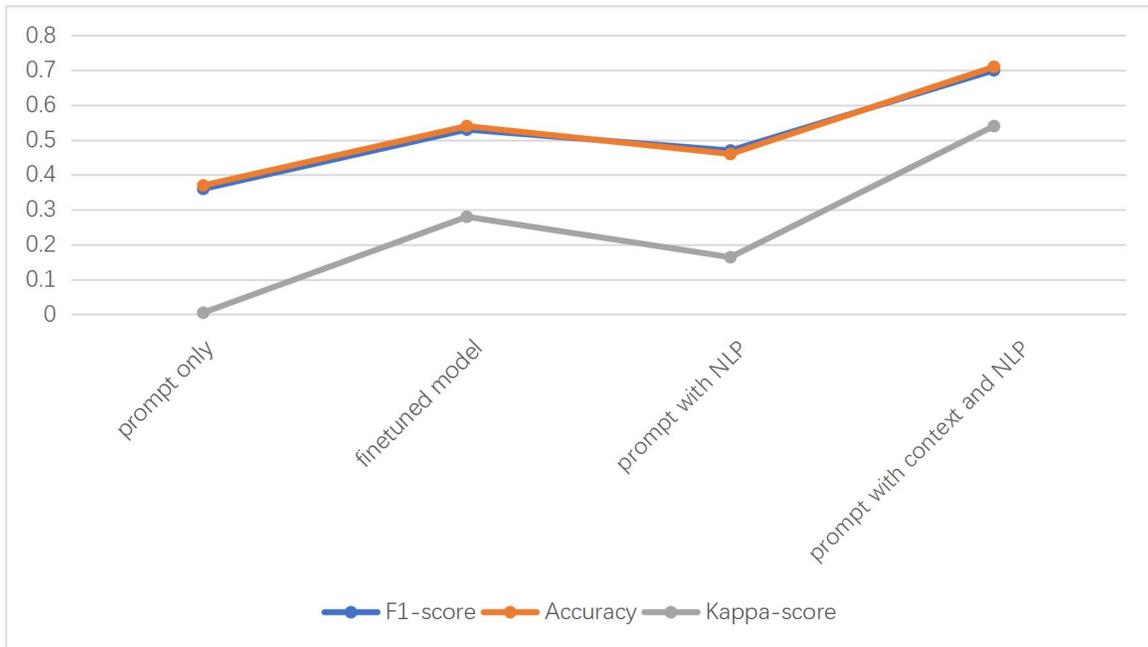

### 5.4. Summary of the results

In this section, we summarized the results of all three automatic deductive coding approaches on annotation and discussion datasets. Because we used several different methods in the GPT-based approach, we selected the best-performed one with context information and NLP integrated. The Kappa and the accuracies of the three approaches are listed in Table 13 and Table 14 respectively.

**Table 13**

*Comparison of Kappa values for three classification methods on two datasets*

|  | Kappa of the annotation classification | Kappa of the discussion classification |
|---|---|---|
| Random Forest | 0.33 | 0.32 |
| RoBERTa | 0.39 | 0.54 |
| GPT4 | 0.55 | 0.72 |

**Table 14**

*Comparison of accuracy values for three classification methods on two datasets*

|  | Accuracy of the annotation classification | Accuracy of the discussion classification |
|---|---|---|
| Random Forest | 0.59 | 0.48 |
| RoBERTa | 0.62 | 0.67 |
| GPT4 | 0.70 | 0.77 |

**RQ1:** Performance of the three classification methods
Randon Forest is the earliest AI algorithm for automatic discourse analysis among the three approaches and GPT is the most cutting-edge one. In general, the more advanced AI approaches produce better performance. Another thing that needs to be noted is that the GPT-based approach, except for the fine-tuned method, only used several labeled samples for writing up the prompt, although we allocated hundreds of labeled training samples for the sake of fair comparison. So, the biggest advantage of the GPT-based approach is not its higher accuracy and kappa, but the low requirement of training samples. This advantage can make the GPT-based approach have value in practice.

**RQ2:** The power of prompt techniques for automatic deductive coding

We need to note that the GPT-based approach is not just plug-and-play for automatic deductive coding. Indeed, ChatGPT provides an easy-to-use interface for all. Even people without any programming experience can use ChatGPT to accomplish some sophisticated tasks like article summarization and data analysis. However, as we showed in the results of the different methods of the GPT-based approach, the performance can be significantly improved when techniques such as RAG and CoT are included because GPT needs contextual information to make more accurate decisions. We also showed that we may need to figure out some rules with traditional NLP techniques like discourse similarity calculation to identify some discourse code, instead of having GPT taking over all the tasks. The reasons behind such integration are probably because we as human experts know more contextual information and guidance than LLMs, and we should describe them as much as possible by using all kinds of LLMs-related techniques such as RAG, CoT, few-shots and so forth.

## 6. Conclusion

In conclusion, this study aimed to assess and compare three distinct text classification methods applied to automatic deductive coding. The methods we adopted encompassed traditional text classification with feature engineering, BERT-like pre-trained language models, and GPT-like pretrained language models, representing generative language models.

The traditional text classification method, employing feature engineering struggled to adapt to the case of student-generated content in both annotation and discussion datasets. On the other hand, the BERT-like model exhibited improved accuracy, leveraging its contextual understanding of language. However, its reliance on tokenized input and the need for substantial data size and computational resources limit its practicality. The standout performer in our study was the GPT-based approach. This approach showcased remarkable adaptability and effectiveness in classifying both annotation and discussion data, outperforming other methods in terms of accuracy and Kappa values. The generative language model, with its inherent ability to consider word order and context, demonstrated promising results even with a limited dataset.

The comparison highlighted the potential of generative language models. The GPT-like model, in particular, presents a compelling avenue for further exploration in educational technology, showcasing promising results in the context of student participatory learning. As the field evolves, leveraging such models could bring more efficient and accurate ways to assess and engage with student-generated content, ultimately enhancing the quality of educational processes.